\documentclass[runningheads]{llncs}

\usepackage{eccv}

\usepackage{eccvabbrv}

\usepackage{graphicx}
\usepackage{booktabs}
\usepackage{amsmath}
\usepackage{booktabs}
\usepackage{xcolor}
\usepackage{tabularx}
\usepackage[table]{xcolor}
\usepackage{wrapfig}
\usepackage{hyperref}

\usepackage{makecell}
\usepackage{booktabs}
\usepackage{colortbl}
\usepackage{graphicx}

\usepackage[accsupp]{axessibility}  

\usepackage{hyperref}

\usepackage{orcidlink}
\usepackage{graphicx}

\begin{document}

\title{\texorpdfstring{$\mathcal{M}^2$}{M2}Tok: Multi-head Multi-codebook Discrete Action Tokenization for Vision-Language-Action Models}

\titlerunning{\texorpdfstring{$\mathcal{M}^2$}{M2}Tok}


\newcommand{\intern}{\textsuperscript{*}}
\newcommand{\corrauth}{\textsuperscript{\textdagger}}

\author{Chunpu Xu\inst{1,3}\intern\orcidlink{0000-0003-4377-9934} \and
Zhixuan Liang\inst{3,5}\corrauth\orcidlink{0009-0008-6815-9866} \and
Yuhao Zhang\inst{2}\orcidlink{0009-0008-3370-9258} \and
Chi-Min Chan\inst{4}\orcidlink{0009-0006-0218-3412} \and
Jiashuo Wang\inst{1}\orcidlink{0000-0002-8254-8138} \and
Yang Xiao\inst{1}\orcidlink{0009-0009-6191-1522} \and
Mengkang Hu\inst{5}\orcidlink{0009-0009-3779-3378} \and
Xiaokang Yang\inst{2}\orcidlink{0000-0003-4029-3322} \and
Yao Mu\inst{2,3}\corrauth\orcidlink{0000-0002-0321-021X}}

\authorrunning{C.~Xu et al.}

\institute{The Hong Kong Polytechnic University, HongKong SAR, China \and
Shanghai Jiao Tong University, Shanghai, China \and
Shanghai AI Laboratory, Shanghai, China \and
Hong Kong University of Science and Technology, HongKong SAR, China \and
The University of Hong Kong, HongKong SAR, China\\
\email{chun-pu.xu@connect.polyu.hk, zxliang@cs.hku.hk, muyao@sjtu.edu.cn}
}

\maketitle

\begingroup
\renewcommand{\thefootnote}{*}
\footnotetext{Work done during internship at Shanghai AI LAB.}
\renewcommand{\thefootnote}{\textdagger}
\footnotetext{Corresponding Authors.}
\endgroup

\begin{abstract}
Recent advancements have successfully adapted autoregressive language models to process multimodal signals, such as images and actions. Since raw action signals are continuous, effective tokenization is essential to map high-dimensional inputs into compact discrete tokens for autoregressive processing. However, existing discrete action tokenizers often suffer from high reconstruction loss, failing to preserve the fine-grained dynamics required for precise control. This ``discretization bottleneck'' significantly limits the performance ceiling of downstream Vision-Language-Action (VLA) models.
To address this, we propose $\mathcal{M}^2$Tok, a Multi-head Multi-codebook Action Tokenizer designed to minimize reconstruction error and enhance policy performance. Our approach introduces two key structural innovations: (1) we decompose the latent action features into multiple heads, enabling the model to implicitly align specific heads with distinct action dimensions; (2) we assign independent codebooks to each head for quantization. By leveraging the combinatorial nature of multiple codebooks, we significantly expand the representational expressivity of the tokenizer, leading to substantially lower reconstruction loss compared to previous methods. 
We evaluate the $\mathcal{M}^2$Tok-based VLA on the RoboTwin, Simpler-Env, and 3 zero-shot real-world tasks. Experimental results demonstrate our method not only achieves superior reconstruction fidelity but also significantly boosts the success rate of VLA models. Comprehensive ablation studies further confirm the effectiveness of the multi-head and multi-codebook mechanisms.  Code is available at \href{https://github.com/cpaaax/M2Tok}{https://github.com/cpaaax/M2Tok}.
\end{abstract}

\section{Introduction}\label{sec:introduction}
Autoregressive language models~\cite{dubey2024llama, liu2024deepseek, achiam2023gpt} have emerged as a unifying paradigm for multimodal generation, demonstrating remarkable capabilities in visual synthesis~\cite{tian2024visual, sun2024autoregressive, DBLP:conf/nips/MaJWYYYPQ25}, audio generation~\cite{ji2025wavtokenizer, zhang2024speechtokenizer}, and robotic control~\cite{brohan2022rt,zitkovich2023rt,kim2024openvla,Seungjae2024vqbet,pertsch2025fast}. By leveraging specialized tokenizers to map continuous signals into compact discrete symbols, Vision-Language-Action (VLA) models can inherit the scalable architectures and training recipes of Large Language Models (LLMs). However, unlike text or image patches, robot actions possess unique characteristics. They are high-frequency, continuous, and exhibit strong inter-step correlations. Designing an action tokenizer that effectively captures these fine-grained dynamics remains a critical bottleneck in current VLA research.

Early approaches, such as RT-1~\cite{brohan2022rt} and OpenVLA~\cite{kim2024openvla}, adopted a naive binning strategy, discretizing each action dimension independently into fixed bins. While simple, this per-step tokenization ignores temporal correlations and struggles with high-frequency control, often leading to jerky motions. To address this, recent works have embraced ``action chunking''~\cite{chi2023diffusion,zhao2023ACT}, predicting sequences of future actions to ensure temporal coherence and mitigate compounding errors. This shift necessitates tokenizers capable of compressing entire action trajectories rather than single steps.

Consequently, two primary paradigms for trajectory tokenization have emerged: frequency-based compression and latent vector quantization. 
FAST~\cite{pertsch2025fast} treats action sequences as signals, utilizing the Discrete Cosine Transform (DCT) combined with Byte-Pair Encoding (BPE)\cite{gage1994new}. While FAST effectively reduces sequence length, it fundamentally clashes with the topology of robotic data that BPE is designed for discrete text, and applying it to continuous signals requires a harsh rounding of DCT coefficients. This introduces the reconstruction loss and results in variable-length tokens, which complicates efficient parallel decoding\cite{kim2025openvla_oft}.
Alternatively, Vector Quantization (VQ) methods, such as VQ-BET~\cite{Seungjae2024vqbet} and VQ-VLA~\cite{Wang2025VQVLA}, learn a fixed-length discrete latent representation via VQ-VAE. However, we identify a critical ``discretization bottleneck'' in these designs, stemming from two fundamental limitations.
First, they suffer from semantic entanglement. Existing tokenizers typically treat the robot action vector as a monolithic entity, projecting heterogeneous signals, such as 6-DoF poses and binary gripper states, into a shared latent space. This approach fails to explicitly model the varying semantics across different action dimensions, forcing the tokenizer to compromise between the high-precision dynamics of arm trajectories and the discrete modes of gripper actuation.
Second, and perhaps more importantly, these methods exhibit limited representational expressivity. By relying on residual vector quantization (RVQ), their capacity to represent diverse behaviors is constrained by the size of the codebook. This structure lacks combinatorial density required to cover the vast, high-frequency action space of manipulation, often resulting in coarse approximations that lose fine-grained control details.

To overcome this bottleneck, we propose $\mathcal{M}^2$Tok, a Multi-head Multi-codebook action tokenizer to maximize representational expressivity and reconstruction fidelity. Our key insight is that high-dimensional action spaces are best represented \begin{wraptable}{r}{0.4\textwidth}
    \centering
    \renewcommand{\arraystretch}{0.4}
    \begin{tabularx}{\linewidth}{>{\centering\arraybackslash}X >{\centering\arraybackslash}X} 
        \toprule
        Tokenizer & L1 loss \\ 
        \midrule
        Fast & 0.0055 \\ 
        VQ-BET & 0.0044 \\ 
        VQ-VLA & 0.0032 \\ 
        $\mathcal{M}^2$Tok & 0.0024 \\ 
        \bottomrule
    \end{tabularx}
        \vspace{-10pt}

    \caption{Reconstruction L1 loss of different tokenizers on RoboTwin data.}
    \vspace{-20pt}
    \label{tab:resconstruct_loss}
\end{wraptable}\noindent not by a residual vector quantization, but by decomposing the latent space into orthogonal subspaces. 
Specifically, $\mathcal{M}^2$Tok introduces two structural innovations that (1) subspace decomposition: we split the latent action features into multiple heads, allowing the model to implicitly align specific heads with distinct underlying dynamics (e.g., end-effector pose vs. gripper status); (2) combinatorial quantization: we assign an independent codebook to each head. Consider a single codebook containing $V$ entries, its representational capacity is consequently $V$. If, however, we partition the available capacity into $h$ equal-sized codebooks, each consisting of $\frac{V}{h}$ entries (such that the total number of code vectors remains $V$), the number of unique action representations becomes $\left( \frac{V}{h} \right)^h$. By utilizing $h$ codebooks of equal size, we exponentially enhance the action feature space's expressivity while maintaining a constant total codebook size. This strategic partitioning dramatically increases the diversity and richness of possible representations, thereby expanding the potential efficacy of the action tokenizer. As shown in Table~\ref{tab:resconstruct_loss}, this architecture allows $\mathcal{M}^2$Tok to achieve reconstruction fidelity significantly superior to existing state-of-the-art tokenizers, breaking the discretization bottleneck. Additionally, building upon the subspace decomposition and combinatorial quantization of 
$\mathcal{M}^2$Tok, we design a simple yet effective conversion module to bridge the gap between multi-codebook tokens and downstream VLA models. This module ensures that the distinct dynamics captured by each head are preserved as coherent semantic units. This design prevents information loss during autoregressive decoding, allowing the VLA model to fully utilize the diverse and rich representation space facilitated by our multi-head multi-codebook architecture.

In summary, our contributions are threefold:
\begin{itemize}
    \item We present $\mathcal{M}^2$Tok, a transformer-based multi-head multi-codebook discrete action tokenizer designed for Vision-Language-Action models. The multi-codebook architecture enhances $\mathcal{M}^2$Tok's ability to represent a wide variety of action sequences. 
    \item We design a lightweight conversion module that enables VLA models to effectively interpret disentangled semantics learned by multi-head multi-codebook architecture, significantly improving downstream VLA performance.
    \item We demonstrate the efficacy of $\mathcal{M}^2$Tok-based VLAs through extensive evaluation on the RoboTwin, Simpler-Env, and three zero-shot real-world tasks, showing that superior reconstruction fidelity directly translates to higher success rates in complex manipulation tasks.
\end{itemize}

\section{Related Work}\label{sec:related}

\subsection{Vision-Language-Action Models}
Vision-Language-Action models (VLAs)~\cite{Shridhar2021CLIPortWA,Shridhar2022PerceiverActorAM,brohan2022rt,zitkovich2023rt,Wu2023GR_1,Cheang2024GR_2,Belkhale2024RTHAH,Team2024OctoAO,liang2025discrete, kim2024openvla,Wang2024ScalingPL,Seungjae2024vqbet,pertsch2025fast} have become a unified framework for robotic manipulation, combining visual perception, language processing, and the generation of robotic actions. These models extend the foundational capabilities of Large Language Models (LLMs) and Vision-Language Models (VLMs) by incorporating action prediction to facilitate seamless interaction with the physical world. Early works, such as CLIPort~\cite{Shridhar2021CLIPortWA} and PerAct~\cite{Shridhar2022PerceiverActorAM}, focused on aligning visual cues with language-driven action policies. The RT series~\cite{brohan2022rt,zitkovich2023rt, Belkhale2024RTHAH} further advanced this field by introducing action tokenization, which enabled scalable transfers from web data to robotic applications. Recent works include Octo~\cite{Team2024OctoAO}, which developed a multi-robot dataset to enhance multitask learning, and OpenVLA~\cite{kim2024openvla}, showcasing significant generalization in executing household tasks. 

\subsection{Action Tokenization}
Bridging continuous control with discrete LLMs hinges on effective action tokenization. While early methods like RT-1~\cite{brohan2022rt} and OpenVLA~\cite{kim2024openvla} employed naive per-step binning, they neglect temporal correlations and high-frequency dynamics. Consequently, recent research has shifted towards trajectory tokenization to encode action chunks.
Frequency-based approaches, such as FAST~\cite{pertsch2025fast}, utilize DCT combined with BPE. However, applying text-centric BPE to continuous signals forces a topological mismatch, introducing reconstruction errors and yielding variable-length tokens that complicate parallel decoding~\cite{kim2025openvla_oft}.
Alternatively, VQ-based methods~\cite{Seungjae2024vqbet,Wang2025VQVLA} map trajectories to discrete latents but suffer from a ``discretization bottleneck'': their reliance on monolithic codebooks entangles heterogeneous semantics (e.g., pose vs. gripper) and constrains expressivity to the size of the vocabulary.
Unlike these approaches, our $\mathcal{M}^2$Tok utilizes a multi-head, multi-codebook architecture to achieve combinatorial expressivity, explicitly disentangling action semantics while maintaining a fixed token length for streamlined autoregressive generation. 
Alternatively, recent works have bypassed tokenization by integrating continuous diffusion heads directly into transformer architectures~\cite{Liu2024RDT,Hu2024VideoPP,Nvidia2025GR00TNA,Intelligence2025pi_0_5,Li2025UnifiedVA,Bu2025UniVLALT,Liu2025HybridVLA}. These approaches require modifying the standard LLM architecture and loss functions (e.g., incorporating denoising objectives). In contrast, our approach retains the pure discrete autoregressive interface, allowing $\mathcal{M}^2$Tok to seamlessly inherit the pre-training recipes and inference optimizations of standard LLMs without architectural deviation.
\section{Method}\label{sec:method}
\begin{figure*}[t]
    \centering
    \includegraphics[width=0.98\textwidth]{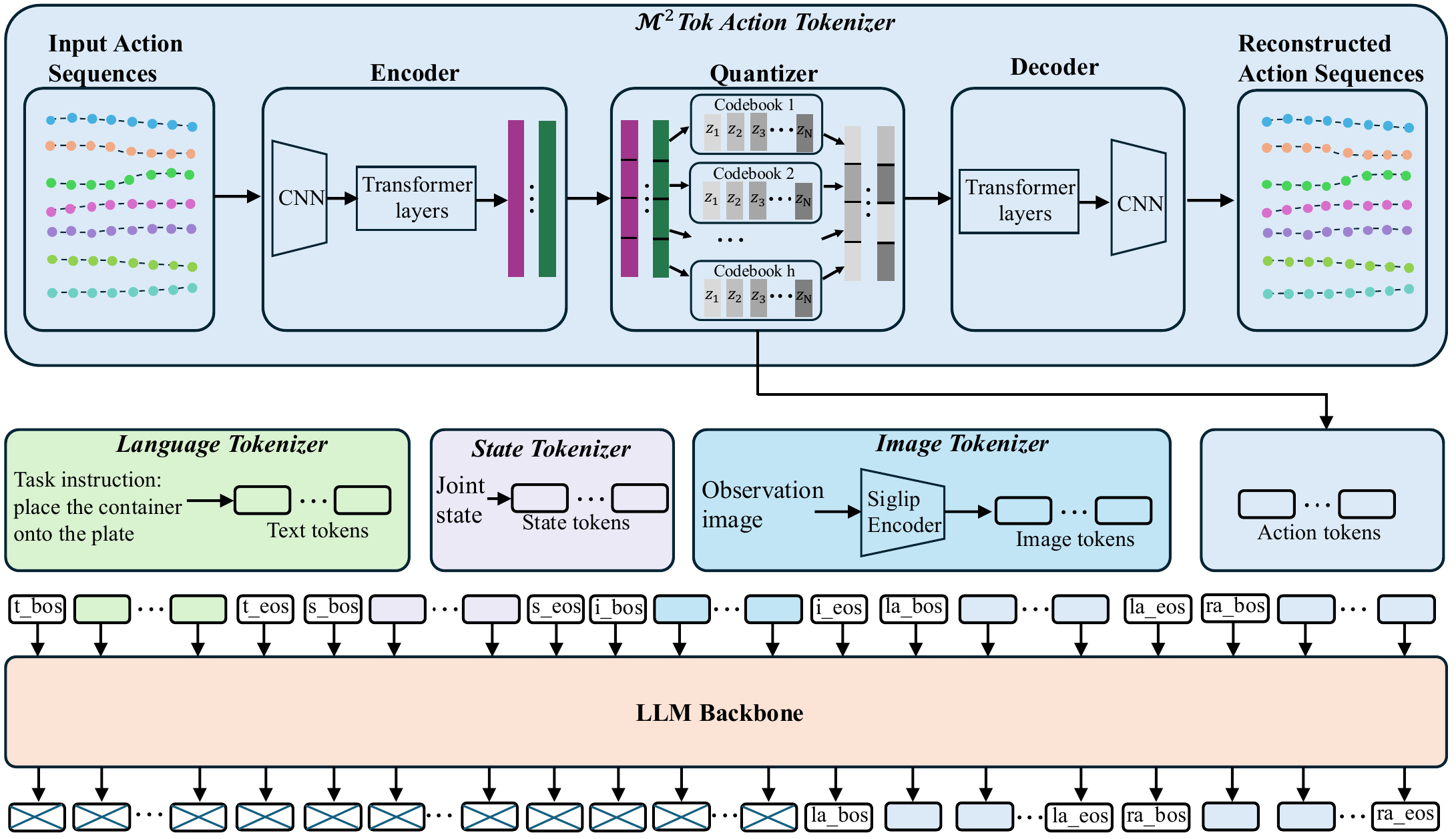}
    \caption{{Overview of $\mathcal{M}^2$Tok action tokenizer (top) and the $\mathcal{M}^2$Tok-based VLA model (bottom).}}
    \label{fig:overview}
    \vspace{-15pt}
\end{figure*}

The overview of the $\mathcal{M}^2$Tok-based VLA is demonstrated in Figure \ref{fig:overview}. The VLA training process involves two key steps. Initially, we train the $\mathcal{M}^2$Tok to function as an action tokenizer. Subsequently, we integrate the $\mathcal{M}^2$Tok with the VLA to facilitate multitask learning.

\subsection{Problem Formulation}
We consider the problem of learning a general-purpose robotic policy from a diverse multi-task datase $D = \{\{(o_{1}, s_{1}, a_{1}),..., (o_{T_i}, s_{T_i}, a_{T_i}), L_{m, i}\}_{i=1}^{N_m}\}^{M}_{m=1}$. Here, $M$ denotes the number of tasks, $N_m$ is the number of trajectories for the $m$-th task, and $L_{m,i}$ provides the natural language description of the $i$-th trajectory in the $m$-th task. Each trajectory consists of a sequence of observations $o_t$ (RGB images), proprioceptive states $s_{t_{m}}$ and continuous bimanual actions $a_t\in \mathbb{R}^{14}$, where each arm possesses 7 Degrees of Freedom (DoF). Our objective is to train a policy $\pi_{\theta}(a_{t:t+k}|o_t,s_t,L)$, which predicts a coherent sequence of future actions $a_{t:t+k}$ (an action chunk of length $k$) given the current observation and instruction. However, modeling the high-dimensional continuous distribution of $a_{t:t+k}$ directly is intractable for standard language models. Therefore, we introduce $\mathcal{M}^2$Tok to discretize the continuous action space into compact tokens, enabling the policy to generate complex behaviors via standard categorical cross-entropy optimization.

\subsection{$\mathcal{M}^2$Tok Tokenizer}
We construct $\mathcal{M}^2$Tok upon the Vector Quantized Variational Autoencoder (VQ-VAE) framework~\cite{van2017neural}, specifically adapted to address the "discretization bottleneck" in robotic control. As illustrated in Figure~\ref{fig:overview}, the architecture comprises three key components: a Temporal-Aware Hybrid Encoder, a Multi-Head Multi-Codebook Quantizer, and a Fidelity-Preserving Decoder. 

To maximize data efficiency and enforce kinematic symmetry, we propose a Bimanual Factorization strategy. Instead of treating the 14-DoF dual-arm action as a monolithic vector, we decompose each bimanual trajectory $a_{1:T}$ into two independent single-arm sequences, $a^{left}$, $a^{right} \in \mathbb{R}^{T\times7}$. These single-arm trajectories form a unified action set $A'$, which serves as the input to our tokenizer. This design encourages the model to learn arm-agnostic kinematic primitives, leaving inter-arm coordination to be handled by the high-level policy.

The encoder is designed to capture both high-frequency local dynamics and long-range temporal dependencies. We employ a hybrid architecture that interleaves 1D causal Convolutional layers with Transformer blocks. The convolutional layers extract local motion features and downsample the temporal dimension, while the Transformer layers integrate global context. Given an input single-arm action chunk $a'\in \mathbb{R}^{k\times 7}$ from $A'$, the encoder produces a continuous latent sequence $\hat{Z} \in \mathbb{R}^{R\times d}$, where $R$ is the reduced temporal resolution and $d$ is the latent dimension.

Standard VQ-VAEs typically quantize the entire latent vector using a single codebook, limiting expressivity to the codebook size $|V|$. To overcome this limitation, we introduce a Combinatorial Quantization mechanism. We decompose the latent space $\mathbb{R}^d$ into $h$ orthogonal subspaces (heads). For each latent vector $\hat{z}_r$ ($r\in\{1,...,R\}$) is split into $h$ segments $\{\hat{z}_{i,r}\}^{h}_{i=1}$, where each segment $\hat{z}_{i,r} \in \mathbb{R}^{\frac{d}{h}}$. Crucially, we maintain $h$ independent codebooks $C=\{Z_1, ...,Z_h\}$, where each codebook $Z_i=\{z_{i,n}\}_{n=1}^N$ contains $N$ learnable prototypes of dimension $\frac{d}{h}$. The quantization is performed independently within each subspace:
\begin{equation}
    q_{i,r} = arg {\underset{z_{i,n}\in Z_i} {min}} ||z_{i,n}-\hat{z}_{i,r} ||_2,
\end{equation}
where $q_{i,r}$ is the nearest neighbor lookup index
from the codebook $Z_i$.

This process effectively expands the representational capacity from $N$ to $(\frac{N}{h})^h$ by combining codes from different subspaces. The quantized latent vector $z_{q,r}$ is then reconstructed by concatenating the selected prototypes: 
\begin{equation}
z_{q,r} = \text{Concat}\left(z_{1, q_{1,r}}, \dots, z_{h, q_{h,r}}\right).
\end{equation}
This multi-head design allows $\mathcal{M}^2$Tok to disentangle distinct semantic attributes of the action (e.g., end-effector pose vs. gripper state) into different subspaces, significantly enhancing reconstruction fidelity. The decoder mirrors the encoder's hybrid architecture. It takes the quantized sequence of latent vectors $Z_q=\{{z_{q,r}}\}_{r=1}^R$ and progressively upsamples it to reconstruct the original action $\hat{a}$.

To train the $\mathcal{M}^2$Tok tokenizer, we follow VQ-VAE~\cite{van2017neural} to employ three different loss functions for optimization. The first is the reconstruction loss $\mathcal{L}_{rec}$ to minimize the difference between
the predicted action sequences $\hat{a}$ and the input action sequences $a'$. The second is the embedding loss $\mathcal{L}_{emb}$, which compute the distance between latent representation $\hat{z}_{i,r}$ and the nearest embedding $z_{i, q_{i,r}}$ to update the embedding space of codebooks. The third is the commitment loss $\mathcal{L}_{com}$, which specifically affects the encoder weights, motivating the encoder's output to remain near the selected codebook vector, thus reducing frequent switching between different code vectors.
The total training losses are optimized as follows:
\begin{equation}
\begin{aligned}
    \mathcal{L} &= \lambda_1 \mathcal{L}_{rec} + \lambda_2 \mathcal{L}_{emb} + \lambda_3 \mathcal{L}_{com} \\
    &= \lambda_1  \| \hat{a} - a' \|^2_2 + \lambda_2 \sum_{i=1}^h \sum_{r=1}^R \|sg(\hat{z}_{i,r})-z_{i, q_{i,r}} \|^2_2 \\ &\quad + \lambda_3 \sum_{i=1}^h \sum_{r=1}^R \|sg(z_{i, q_{i,r}})-\hat{z}_{i,r} \|^2_2,
\end{aligned}
\end{equation}
where $\lambda_1$, $\lambda_2$ and $\lambda_3$ are the weights to each loss item, $\textit{sg}$ represents stopgradient operation.

\subsection{$\mathcal{M}^2$Tok-based VLA}
Our goal is to train a VLA policy $\pi_{\theta}(a_{t:t+k}|o_t,s_t,L)$.
For the language description \(L\), we use the base LLM’s native tokenizer to obtain text tokens. Regarding the joint states $s_t$, we adopt the approach outlined in FAST~\cite{pertsch2025fast} to discretize each joint dimension into 256 bins, yielding a sequence of discrete state tokens. For processing the observation image $o_t$, we resize each image to a resolution of $224 \times 224$ and then apply siglip-so400m-patch14-224~\cite{Zhai2023Siglip} as the image tokenizer. This procedure produces a 16×16 grid of patch embeddings, i.e., 256 continuous image tokens per frame. For the input action sequence, we use the proposed $\mathcal{M}^2$Tok to transform the continuous action sequences $a'$ into discrete action tokens $\tilde{a}$. To delineate modality boundaries in the input tokens, we introduce special functional tokens: \textit{t\_bos}/\textit{t\_eos} for text, \textit{s\_bos}/\textit{s\_eos} for state, \textit{i\_bos}/\textit{i\_eos} for image, and \textit{la\_bos}/\textit{la\_eos} and \textit{ra\_bos}/\textit{ra\_eos} for left-arm and \textit{right-arm actions}, respectively. During training, we serialize the context and targets into a single sequence, for example:
[\textit{t\_bos}, \textit{t\_tokens}, \textit{t\_eos}, \textit{s\_bos}, \textit{s\_tokens}, \textit{s\_eos}, \textit{i\_bos}, \textit{i\_tokens}, \textit{i\_eos}, \textit{la\_bos}, \textit{left-action\_tokens}, \textit{la\_eos}, \textit{ra\_bos}, \textit{right-action\_tokens}, \textit{ra\_eos}]. This design yields a consistent tokenized interface across modalities, enabling straightforward autoregressive learning of the action sequence conditioned on language, state, and vision.

Our $\mathcal{M}^2$Tok converts continuous actions into fixed-length sequences of discrete tokens. We explore the effectiveness of the proposed action tokenizer by employing the autoregressive decoding.

\begin{wrapfigure}{r}{0.5\textwidth}
\vspace{-10pt}
    \centering
    \includegraphics[width=0.95\linewidth]{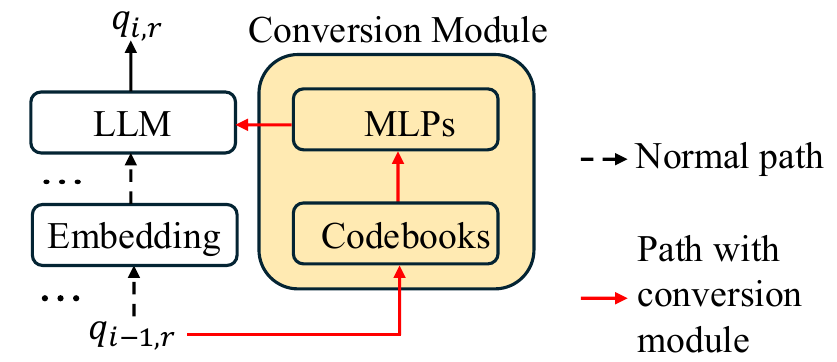}
        \caption{\textbf{$\mathcal{M}^2$Tok conversion module.}}
        \label{fig:conversion}
        \vspace{-10pt}
\end{wrapfigure}

\noindent\textbf{Autoregressive Decoding.}
In the autoregressive setting, the VLA generates the action sequence via next-token prediction. At each position of generation, the model outputs a distribution over the action vocabulary conditioned on the input context $(o_t, s_t,L)$ and all previously generated action tokens. The training objective is to minimize the negative log-likelihood of the predicted action tokens, given the sequence of preceding tokens:
\vspace{-2pt}
\begin{equation}
    \mathcal{L}_{AD} = -\sum_{i=1}^h \sum_{r=1}^R \text{log} P(q_{i,r}|q_{<}, o_t,s_t,L),
\end{equation}
\vspace{-2pt}
where $q_{<}$ represent previous action tokens.

In the standard setup, action tokens are handled like text tokens: the previous discrete action token $q_{i-1,r}$ is first mapped to a vector by the token-embedding layer, and the LLM then uses this representation to predict the next token $q_{i,r}$. To better leverage the action semantics captured by $\mathcal{M}^2$Tok, we replace this generic embedding with a lightweight action-token conversion module. As shown in Figure \ref{fig:conversion},  this module maps each discrete action token to a continuous representation that preserves the semantics learned by $\mathcal{M}^2$Tok, providing the VLA with more informative action features while keeping the rest of the generation pipeline unchanged. Concretely, for each previously generated action token $q_{i-1,r}$, we retrieve its corresponding quantized embedding $z_{i-1, q_{i-1,r}}$ from the $\mathcal{M}^2$Tok multi-codebooks. We then pass the quantized embedding through an MLP projector that maps it into the same semantic space as the LLM's text embeddings.

\section{Experiments}\label{sec:experiments}

In Section \ref{subsec:implementation}, we describe the implementation details of our experiments. Section \ref{subsec:simulation} evaluates the manipulation performance of the $\mathcal{M}^2$-based VLA in the RoboTwin and Simpler-Env simulation, comparing it against prior methods. In Section \ref{subsec:ablation}, we present ablations that quantify the contribution of each component. Section \ref{subsec:quantitative} analyzes how the number of codebooks influences VLAs' performance. In Section  \ref{subsec:latency}, we analyze the latency of VLAs based on different tokenizer. Finally, Section \ref{subsec:real-world} reports both quantitative and qualitative results on real-world manipulation tasks.

\begin{table}[t]
    \caption{The average number of steps and description for each task of RoboTwin.}
        \vspace{-5pt}

    \resizebox{0.98\linewidth}{!}{%
    \centering
    \begin{tabular}{
    >{\centering\arraybackslash}m{2.7cm} >{\centering\arraybackslash}m{1.2cm} >    {\centering\arraybackslash}m{4cm}|
    >{\centering\arraybackslash}m{2.7cm} >{\centering\arraybackslash}m{1.2cm} >    {\centering\arraybackslash}m{4cm}
    }
        \toprule
        \textbf{Task Name} & \textbf{Avg. Step} & \textbf{Task Example Description} & \textbf{Task Name} & \textbf{Avg. Step} & \textbf{Task Example Description} \\
        \midrule
        Beat Block Hammer & 67 & Grab the silver hammer and use it to hit. & Pick Diverse Bottles & 75 & Grab the plastic bottle, pick up the yellow body bottle. \\ 
        Click Bell & 52 & Press the center top of the blue bell. & Place Mouse Pad & 88 & Grab the gray mouse and drop it on the black mat. \\ 
        Handover Mic & 134 & Grasp the blue microphone and pass it across. & Shake Bottle & 133 & Shake the orange bottle after lifting it. \\ 
        Move Can Pot & 90 & Lift the smooth sauce can, place it by the kitchen pot. & Place Phone Stand & 82 & Carry the flat phone to the green phone rack. \\ 
        Move Pillbottle Pad & 88 & Hold the white bottle and position it on the pad. & Place Burger Fries & 141 & Move the hamburg and fries box to the tray. \\ 
        Move Playingcard Away & 70 & Slide the box for playing cards off the table outward. & Place Container Plate & 92 & Move the cup and drop the container onto the plate.  \\ 
        \bottomrule
    \end{tabular}}
        \vspace{-15pt}

    \label{tab:task_steps}
\end{table}

\subsection{Implementation Details}\label{subsec:implementation}

\noindent\textbf{RoboTwin Simulation.}
We train the VLAs on a synthetic dataset generated with the RoboTwin simulator~\cite{Chen2025RoboTwin}, which enables large-scale, automated collection of diverse and realistic bimanual manipulation trajectories. The dataset spans 12 tasks, each with 1,600 expert demonstrations (19,200 trajectories in total). To increase diversity and robustness, we apply domain randomization during data collection; the specific settings are listed in Table \ref{tab:domain_randomization_setting}. The details of each task are shown in Table~\ref{tab:task_steps}. Observations consist of RGB images from head camera.

\noindent\textbf{Simpler-Env Simulation.}
SimplerEnv bridges the control and visual disparities inherent in simulation, facilitating policy evaluations that accurately predict real-world performance. In this study, we utilize the WidowX robot to demonstrate this capability across four manipulation tasks: Put Carrot on Plate, Put Spoon on Towel, Stack Green on Yellow, and Put Eggplant in Basket. Following prior works, we employ the BridgeV2 dataset~\cite{walke2023bridgedata} as the training data, which is a large-scale dataset featuring 24 distinct environments (e.g., kitchens, sinks, tabletops), over 100 objects, and diverse manipulation tasks.

\begin{table}[t]
\caption{Settings for Data Collection in RoboTwin Simulator.}
\vspace{-5pt}
\centering
\resizebox{0.8\textwidth}{!}{
\begin{tabular}{l c | l c}
\toprule
\textbf{Parameter} & \textbf{Value} &\textbf{Parameter} & \textbf{Value} \\ 
\midrule
        Save Frequency & 15 & Random Head Camera Distance & 0.03 \\ 
        Embodiment & Piper & Random Table Height & 0.03 \\ 
        Random Background & True & Random Light & True \\ 
        Cluttered Table & False & Crazy Random Light Rate & 0.02 \\ 
        Clean Background Rate & 0.02 & Head Camera Type & D435 \\ 
\bottomrule

\end{tabular}
}
\label{tab:domain_randomization_setting}
\vspace{-10pt}
\end{table}

\noindent\textbf{Training setup.} To ensure a fair comparison across action tokenizers, we integrate each tokenizer into the same VLA architecture, as shown at the bottom of Figure \ref{fig:overview}. We use Qwen2.5-0.5B~\cite{Yang2024Qwen2_5} as the LLM backbone. For $\mathcal{M}^2$Tok, the total codebook size is 2048, partitioned into 8 codebooks ($h = 8$) with 256 entries each, and a codebook embedding dimension of 256. The per-arm action dimension is $H = 7$, yielding 14 dimensions for the RoboTwin bimanual setting and Simpler-Env unimanual setting. $\mathcal{M}^2$Tok is trained with AdamW~\cite{Loshchilov2017DecoupledWD} at a learning rate of $5 \times 10^{-5}$, using a total batch size of 2048$\times$4 on four 4090 GPUs.

Subsequent to pre-training the $\mathcal{M}^2$Tok tokenizer, we leverage it to discretize continuous action trajectories into latent tokens for VLA training. During this phase, the parameters of both the  $\mathcal{M}^2$Tok and the visual encoder (SigLIP) are kept frozen, exclusively updating the weights of the LLM backbone and the projection MLPs. We employ the AdamW optimizer with a base learning rate of $1\times 10^{-4}$, modulated by a cosine decay scheduler. The models are trained for 10 epochs on the RoboTwin benchmark and 20 epochs on Simpler-Env to ensure convergence. Additionally, a warm-up ratio of 0.03 is applied to learning rate.

\noindent\textbf{Baselines.} In order to assess the effectiveness of the proposed $\mathcal{M}^2$Tok, we compare it against four baseline tokenizers. Binning tokenizer~\cite{kim2024openvla}: this tokenizer converts the action at each timestep across each action dimension into 256 discrete bins. FAST tokenizer~\cite{pertsch2025fast}: the tokenizer employs DCT to convert action sequences into the frequency domain. It then uses BPE to encode and compress the action tokens. VQ-BET tokenizer~\cite{Seungjae2024vqbet}: VQ-BET uses the designed encoder to encode input action chunks, followed by a two-layer RVQ process to quantize the processed features.  VQ-VLA Tokenizer~\cite{Wang2025VQVLA}: The VQ-VLA utilizes 2D temporal convolutional layers, combined with time and action-type embeddings, to process input actions. These are subsequently passed through RVQ layers to generate discrete action tokens.

To ensure a fair comparison, we retrain the FAST, VQ-BET, and VQ-VLA tokenizers using the same action sequence training dataset employed by $\mathcal{M}^2$Tok. All VLAs are trained in the multi-task setting.

\begin{table*}[t]

\caption{Success rates of different VLAs across 12 RoboTwin tasks.}
\vspace{-5pt}
\centering
\resizebox{0.98\linewidth}{!}{%
    \begin{tabular}{c| >{\centering\arraybackslash}m{1.8cm} >{\centering\arraybackslash}m{2.5cm} >{\centering\arraybackslash}m{1.8cm} >{\centering\arraybackslash}m{1.6cm} >{\centering\arraybackslash}m{2cm} >
    {\centering\arraybackslash}m{1.0cm} >{\centering\arraybackslash}m{1.7cm}}  
        \toprule
        \textbf{Tokenizer} & \textbf{Beat Block Hammer} & \textbf{Move Playingcard  Away} & \textbf{Pick Diverse  Bottles} & \textbf{Move Can Pot} & \textbf{Move Pillbottle  Pad} & \textbf{Click Bell} & \textbf{Handover Mic} \\ 
        \midrule
        \textbf{Binning} &0.11& 0.01& \textbf{0.27}& 0.02&0.00&0.67 &0.85  \\ 
        \textbf{FAST} &0.00& 0.18& 0.06& 0.08& 0.01& 0.68& 0.17 \\ 
        \textbf{VQ-BET} &0.10& 0.27& 0.12& 0.13& 0.06& 0.64& 0.59 \\ 
        \textbf{VQ-VLA} &0.14& 0.46& 0.24& 0.30& 0.20& 0.59& 0.93   \\ 
        \midrule

        \rowcolor{gray!20} 
        \textbf{$\mathcal{M}^2$Tok} &\textbf{0.20}& \textbf{0.48} & 0.22& \textbf{0.58}& \textbf{0.33}& \textbf{0.71}& \textbf{0.94} \\[-0.32em]

        \bottomrule
        \noalign{\vskip 8pt}
        
        \toprule
        \textbf{Tokenizer}& \textbf{Place Mouse  Pad} & \textbf{Place Container Plate} & \textbf{Place Phone  Stand} & \textbf{Place Burger  Fries} & \textbf{Shake Bottle} & \multicolumn{2}{|c}{\textbf{Average Success}} \\ 
        \midrule
        \textbf{Binning} &0.00&0.00&0.02&0.01& 0.92&  \multicolumn{2}{|c}{0.24}\\ 
        \textbf{FAST} &0.01& 0.07& 0.01& 0.00& 0.76&  \multicolumn{2}{|c}{0.17} \\ 
        \textbf{VQ-BET} &0.00& 0.54& 0.04& 0.15& 0.87&  \multicolumn{2}{|c}{0.29} \\ 
        \textbf{VQ-VLA} & 0.08& 0.79& \textbf{0.23}& 0.52& \textbf{0.96} & \multicolumn{2}{c}{0.45}  \\ 

        \midrule

        \rowcolor{gray!20} 
        \textbf{$\mathcal{M}^2$Tok} &\textbf{0.10}& \textbf{0.83} & 0.20& \textbf{0.64}& 0.87 &  \multicolumn{2}{|c}{\textbf{0.51}} \\[-0.32em]
        \bottomrule
    \end{tabular}
    }
\label{tab:main_results}
\vspace{-15pt}
\end{table*}

\subsection{Simulation Results}\label{subsec:simulation}
\begin{figure*}[t]
    \centering
    \includegraphics[width=0.98\textwidth]{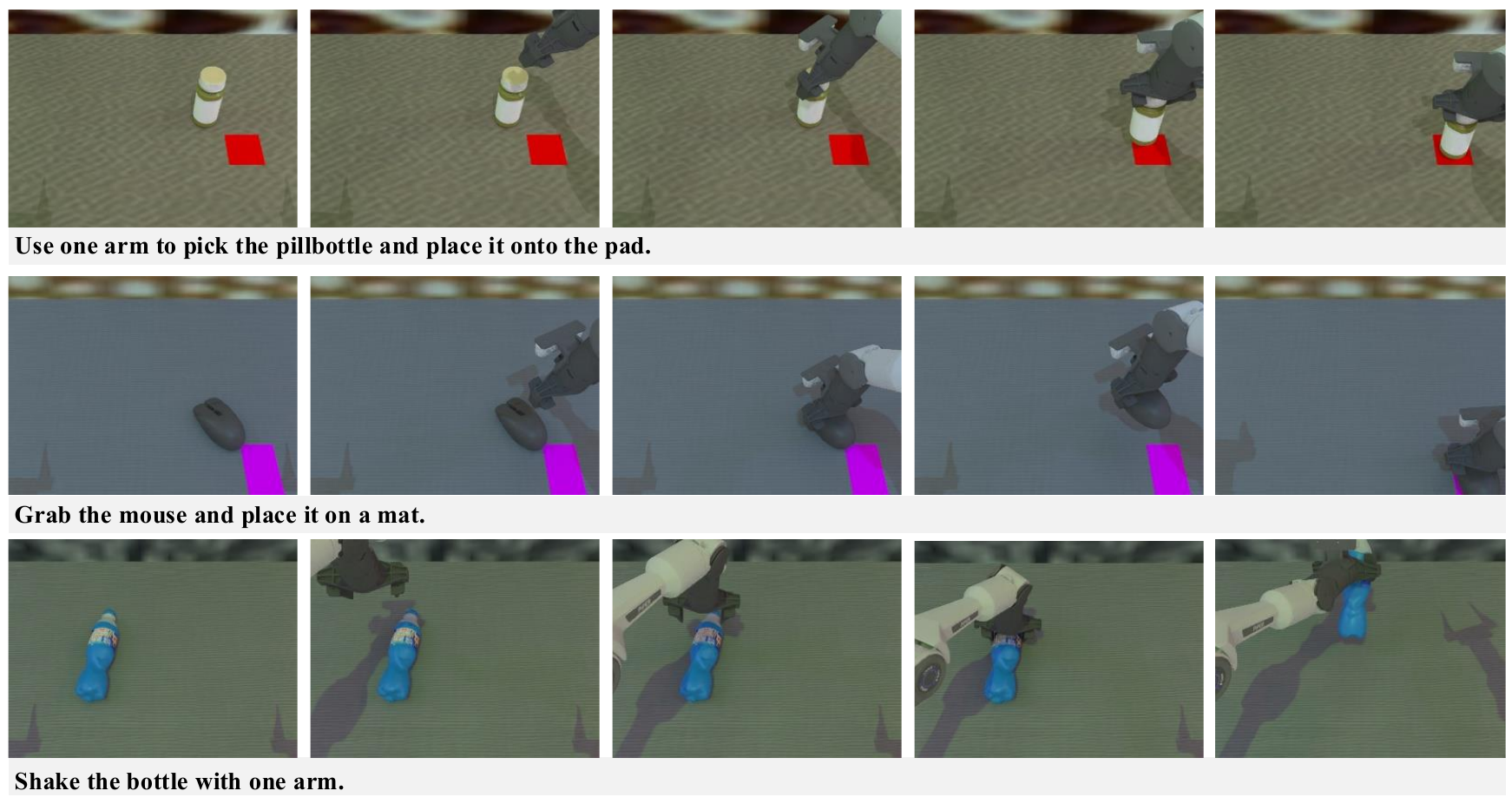}
    \vspace{-5pt}
    \caption{{RoboTwin visualizations for ``Move Pillbottle Pad'', ``Place Mouse Pad'', and ``Shake Bottle'' tasks.}}
    \label{fig:robotwin_examples}
\end{figure*}

\noindent\textbf{RoboTwin.} We evaluate all VLAs on the 12 tasks of RoboTwin outlined in Section \ref{subsec:implementation}. 
We perform 100 rollouts for each task.
Each rollout is labeled as a success (1) or failure (0), and we report the success rate, the fraction of successful rollouts, as the evaluation metric.
Quantitative comparisons across 12 diverse manipulation tasks are presented in Table~\ref{tab:main_results}. Our proposed $\mathcal{M}^2$Tok demonstrates superior efficacy, achieving a state-of-the-art average success rate of {51\%}, surpassing the strongest baseline VQ-VLA (45\%) by a significant relative margin of {13\%}. Notably, in the {Move Can Pot} task, $\mathcal{M}^2$Tok achieves a success rate of {0.58}, nearly doubling the performance of VQ-VLA (0.30). Similarly, in  {Place
Burger Fries}, our method reaches {0.64}, outperforming VQ-VLA (0.52) and dominating VQ-BET (0.12). We attribute this to the {subspace decomposition} mechanism, which disentangles distinct kinematic features (e.g., separating gripper actuation from arm trajectory) into orthogonal heads. This ensures that the VLA model can attend to and predict fine-grained action details necessary for delicate manipulation. Figure \ref{fig:robotwin_examples} presents visualization examples of three tasks within the RoboTwin simulator.

\noindent\textbf{Simpler-Env.} We also conduct experiments on the four tasks of the Simpler-Env benchmark. As summarized in Table~\ref{tab:simpler_results}, $\mathcal{M}^2$Tok demonstrates superior generalization capabilities, achieving the highest average success rate of 28\%, outperforming the strongest VQ-VLA baseline (21\%) by a relative margin of 33\%. 

\begin{figure*}[t]
    \centering
    \includegraphics[width=0.98\textwidth]{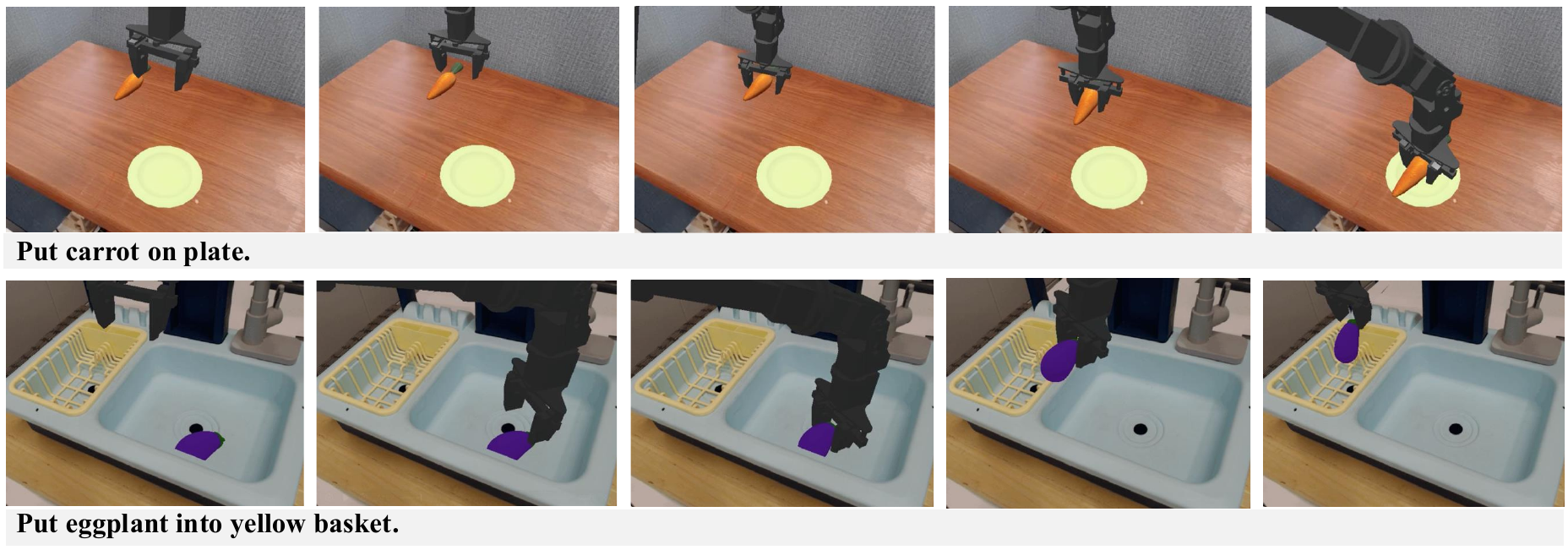}
    \vspace{-5pt}
    \caption{{Simpler-Env visualizations for ``Put Carrot'' and ``Put Eggplant'' tasks.}}
    \label{fig:simpler_env_examples}
    \vspace{-15pt}
\end{figure*}

\begin{wraptable}{r}{0.55\textwidth} 
    \setlength{\tabcolsep}{1.5pt}
    \caption{Success rates of different VLAs across 4 Simpler-Env tasks.}
    \centering
    \resizebox{\linewidth}{!}{ 
        \begin{tabular}{c|c|c|c|c|c} 
        \toprule
        \textbf{Tokenizer} & \textbf{\makecell{Put Spoon \\ on Towel}} & \textbf{\makecell{Put Carrot\\ on Plate}} & \textbf{\makecell{Stack Green \\on Yellow}} & \textbf{\makecell{Put Eggplant\\ in Basket}} & \textbf{Avg.} \\ 
        \midrule
        \textbf{Bining} &0.08   &0.00   &0.00   &0.04   &0.03   \\
        \textbf{FAST} &0.21   &0.17   &0.00   &0.08   &0.12   \\
        \textbf{VQ-BET} &0.04   &0.04   &0.00   &0.00   &0.02  \\
        \textbf{VQ-VLA} &0.29   &0.33   &\textbf{0.21}   &0.00   &0.21  \\
        \midrule
        \rowcolor{gray!20} 
        \textbf{$\mathcal{M}^2$Tok} &\textbf{0.33}   &\textbf{0.38}   &{0.08}   &\textbf{0.33}   &\textbf{0.28} \\
        \bottomrule
    \end{tabular}
    }
    \label{tab:simpler_results}
    \vspace{-15pt} 
\end{wraptable} 
\noindent The most significant performance gain is observed in the {Put Eggplant in Basket} task. While VQ-VLA and other baselines completely failed to solve this task, $\mathcal{M}^2$Tok achieved a remarkable 33\% success rate. This task requires precise grasping of an irregular object (eggplant) and accurate placement. The success of  $\mathcal{M}^2$Tok suggests that our Combinatorial Quantization strategy enables the policy to generate high-fidelity action tokens that can effectively manipulate objects with complex geometries, whereas RVQ codebooks (like VQ-BET) struggle to capture the necessary fine-grained control primitives. Figure \ref{fig:simpler_env_examples} illustrates the visualization results of two tasks in the Simpler-Env simulator.

\begin{table*}[t]
\caption{Results of Ablation study across 12 RoboTwin tasks.}
\centering
\resizebox{0.97\textwidth}{!}{%
    \begin{tabular}{c| >{\centering\arraybackslash}m{1.8cm} >{\centering\arraybackslash}m{2.5cm} >{\centering\arraybackslash}m{1.8cm} >{\centering\arraybackslash}m{1.6cm} >{\centering\arraybackslash}m{2cm} >
    {\centering\arraybackslash}m{1.0cm} >{\centering\arraybackslash}m{1.7cm}} 
        \toprule
        \textbf{Tokenizer} & \textbf{Beat Block Hammer} & \textbf{Move Playingcard  Away} & \textbf{Pick Diverse  Bottles} & \textbf{Move Can Pot} & \textbf{Move Pillbottle  Pad} & \textbf{Click Bell} & \textbf{Handover Mic} \\ 
        \midrule
        \textbf{$\mathcal{M}^2$Tok} &\textbf{0.20}& \textbf{0.48}& \textbf{0.22}& \textbf{0.58} & \textbf{0.33}& \textbf{0.71}& \textbf{0.94} \\ 
        \textbf{w/o Multi-head} &0.10& 0.25& 0.03& 0.12& 0.00& 0.32& 0.59 \\ 
        \textbf{w/o Multi-codebook} &0.14& 0.34& 0.18& 0.34& 0.09& 0.55& 0.65 \\ 
        \textbf{w/o Conversion} &0.13& 0.35& 0.20& 0.35& 0.15& 0.53& 0.81 \\ 
        \bottomrule
        \noalign{\vskip 5pt}
        \toprule
        \textbf{Tokenizer}& \textbf{Place Mouse  Pad} & \textbf{Place Container  Plate} & \textbf{Place Phone  Stand} & \textbf{Place Burger  Fries} & \textbf{Shake Bottle} & \multicolumn{2}{|c}{\textbf{Average Success}} \\ 
        \midrule
        \textbf{$\mathcal{M}^2$Tok} &\textbf{0.10}& \textbf{0.83}& \textbf{0.20}& 0.64 & 0.87&  \multicolumn{2}{|c}{\textbf{0.51}} \\ 
        \textbf{w/o Multi-head} &0.00& 0.36& 0.01& 0.12& 0.79&  \multicolumn{2}{|c}{0.22} \\ 
        \textbf{w/o Multi-codebook} &0.01& 0.46& 0.07& 0.46& 0.84&  \multicolumn{2}{|c}{0.34} \\ 
        
        \textbf{w/o Conversion} &0.07& 0.79& 0.09& \textbf{0.67}& \textbf{0.88}&  \multicolumn{2}{|c}{0.42} \\ 
        \bottomrule
    \end{tabular}
}
\label{tab:ablation}

\end{table*}

\begin{figure*}[t]
    \centering
    \includegraphics[width=0.98\textwidth]{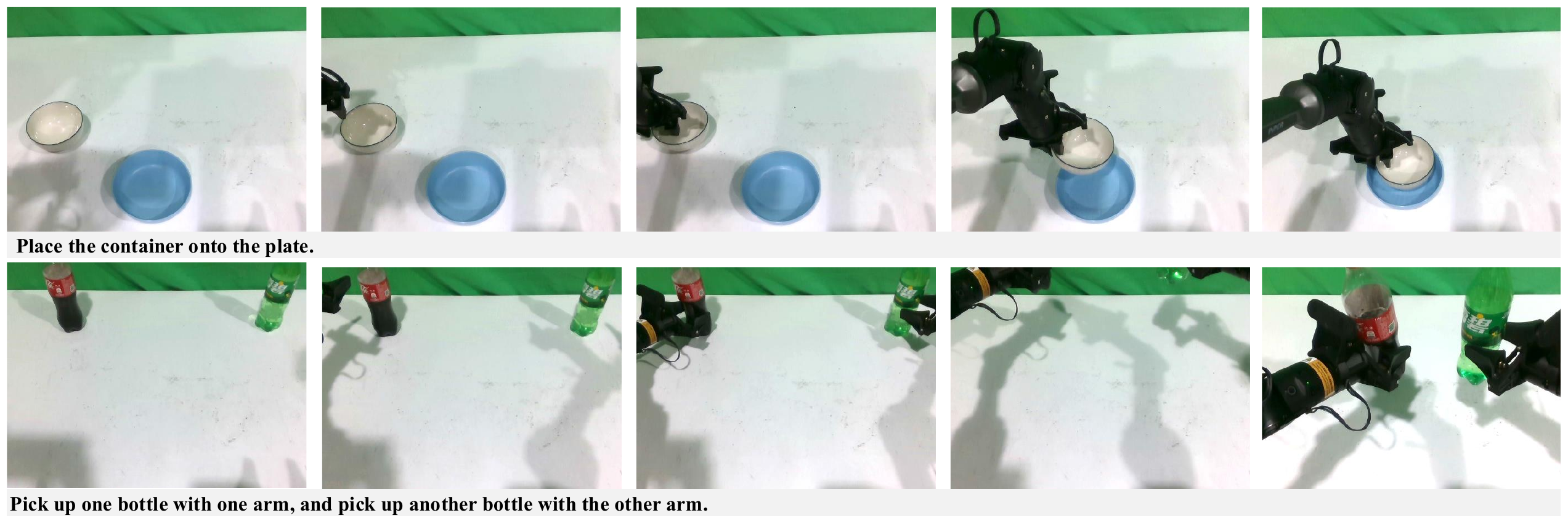}
    \caption{{Real-world manipulation visualizations for ``Place Container Plate'' and ``Pick Diverse Bottles'' tasks.}}
    \label{fig:real_robot}
\end{figure*}

\subsection{Ablation Study}\label{subsec:ablation}
To further investigate the effectiveness of the multi-head architecture, multi-codebook quantization, and the conversion module in $\mathcal{M}^2$Tok, we conducted ablation experiments by individually removing each component while maintaining the others. The experimental setups are as follows:
(1) w/o Multi-head: replacing the multi-head mechanism with a single head to evaluate the importance of subspace decomposition.
(2) w/o Multi-codebook: utilizing a single codebook instead of multiple quantization to assess the impact of combinatorial expressivity.
(3) w/o Conversion: removing the conversion module and directly using the embedding layer to process discrete tokens. 

The results are summarized in Table~\ref{tab:ablation}. We observe that the full $\mathcal{M}^2$Tok model achieves the best performance (51\%), and removing any component leads to a degradation in the overall success rate.  Removing the multi-head mechanism causes the most catastrophic performance drop, plummeting the success rate from 51\% to 22\%. This degradation is particularly severe in precision-dependent tasks like Move Pillbottle Pad (0.33 to 0.00) and Place Mouse Pad (0.10 to 0.00). This result highlights the critical role of kinematic disentanglement. Without multiple heads to independently model different action subspaces (e.g., separating gripper actuation from arm translation), the tokenizer fails to capture the fine-grained dynamics required for complex manipulation, leading to a ``smearing'' of action details. Replacing the diverse codebooks with a single shared codebook results in a significant decline to 34\%. This indicates that the {expressive capacity} of the tokenizer is heavily reliant on the combinatorial nature of our design. A single codebook creates a bottleneck, limiting the diversity of representable action primitives. Removing the conversion module leads to a moderate decrease in performance (0.51 to 0.42). While the model retains some capability, the drop indicates that the conversion module plays a vital role in aligning the discrete latent space of the VQ-VAE with the continuous embedding space of the LLM. It acts as a bridge that smoothes the transition from quantized tokens to autoregressive prediction, thereby enhancing the overall stability and accuracy of the VLA policy.

\begin{figure*}[t]
    \centering
    \includegraphics[width=0.98\textwidth]{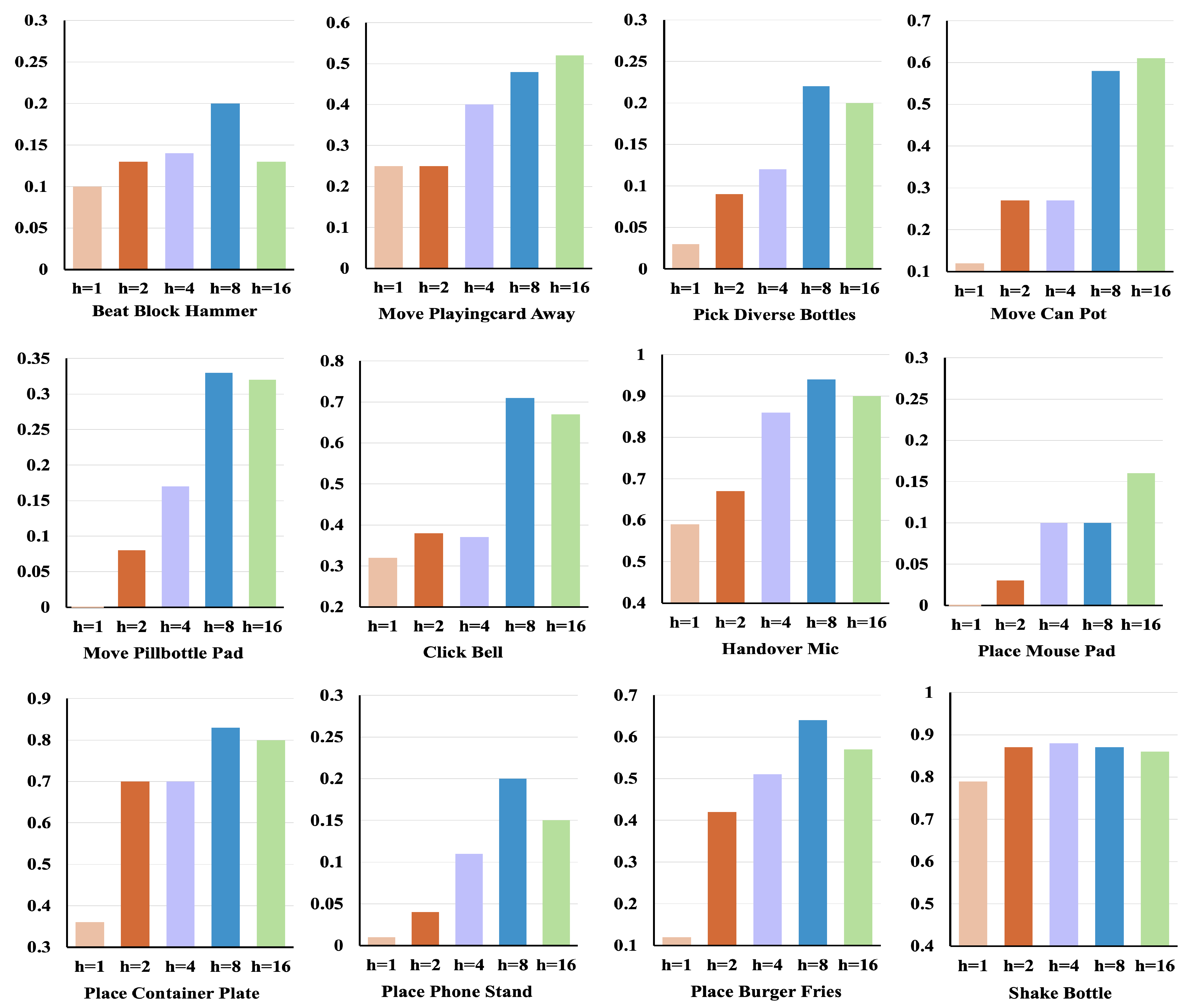}
    \vspace{-10pt}
    \caption{{Effect of varying number $h$ across 12 RoboTwin tasks.}}
    \label{fig:multi-codebook}
    \vspace{-7pt}
\end{figure*}

\subsection{Analysis on the number of Multi-head Multi-codebook}\label{subsec:quantitative}

Recall that the latent features from the $\mathcal{M}^2$Tok encoder are split into $h$ segments, each of which is then quantized by a independant codebook. To quantitatively explore this, we evaluate VLA performance across varying the number of $h$, as shown in Figure~\ref{fig:multi-codebook}. The results align with our theoretical analysis: increasing $h$ significantly boosts performance, confirming that the expanded subspace decomposition and combinatorial diversity allow the tokenizer to capture more nuanced manipulation behaviors without increasing the total vocabulary size. We observe that performance improves consistently for most tasks as $h$ increases, peaking at $h=8$. Further increasing $h$ yields diminishing returns, likely due to the over-fragmentation of the latent space.

\begin{table}[t]
\centering
\small
\renewcommand\arraystretch{0.25}

  \caption{Latency and Frequency in the RoboTwin Environment. }
\vspace{-10pt}
\resizebox{0.7\linewidth}{!}{%
\begin{tabular}{c |c|c|c|c|c|c}
\toprule
\textbf{Method} & \textbf{Bining} & \textbf{FAST} & \textbf{VQ-BET} & \textbf{VQVLA} & \textbf{$\mathcal{M}^2$Tok} & \textbf{$\mathcal{M}^2$Tok(vLLM)}\\
\midrule
\textbf{Latency(ms)} &485.3  &42.7  & 30.7  & 81.2 & 121.6 & 17.8 \\
\midrule
\textbf{Speed(Hz)} &2.0  &23.4 &32.6  &12.3 &8.2 & 56.2 \\

\bottomrule
\end{tabular}}

    \label{tab:Latency}
    \vspace{-15pt}

\end{table}

\subsection{Latency Analysis}\label{subsec:latency}
We evaluate the inference frequency and latency of VLAs based on different tokenizers on a single  RTX 4090 in the bimanual RoboTwin environment, with results shown in Table~\ref{tab:Latency}. Bining-based VLA runs at 2 Hz, while our method increases the frequency to over 8 Hz—a 4.1× improvement. Furthermore, our VLA supports the vLLM inference engine. By leveraging vLLM, our VLA achieves a throughput of 56.2 Hz. This significantly exceeds the control frequency required for most real-time manipulation tasks.

\begin{wraptable}{r}{0.55\textwidth} 
    \setlength{\tabcolsep}{1.5pt}    
    \vspace{-30pt}                   
    \caption{The results of real-world experiments.}
    \centering
    \resizebox{\linewidth}{!}{       
        \begin{tabular}{c|c|c|c|c}   
        \toprule
        \textbf{Tokenizer} & 
        \makecell{\textbf{Click} \\ \textbf{Bell}} & 
        \makecell{\textbf{Place Container}\\ \textbf{Plate}} & 
        \makecell{\textbf{Pick Diverse} \\ \textbf{Bottles}} & 
        \makecell{\textbf{Average}\\ \textbf{Success}} \\ 
        \midrule
        \textbf{Binning} &6/20  &0/20  &0/20  & 0.10\\ 
        \textbf{FAST} &4/20  &1/20  &0/20  &0.08 \\ 
        \textbf{VQ-BET} &7/20  &2/20  &0/20  &0.15  \\ 
        \textbf{VQ-VLA} &10/20  &7/20  &0/20  &0.28  \\ 
        \midrule
        \rowcolor{gray!20} 
        \textbf{$\mathcal{M}^2$Tok} &11/20  &7/20  &2/20  &0.33 \\ 
        \bottomrule
    \end{tabular}
    }
    \label{tab:real_world}
    \vspace{-15pt} 
\end{wraptable}

\subsection{Real-World Results}\label{subsec:real-world}
In our real-world experiments, we utilize the AgileX Cobot Magic, a mobile platform configured with an Aloha setup that includes four robotic arms. Each arm is an AgileX Piper featuring six degrees of freedom and is equipped with a one-DoF parallel gripper. The platform is also outfitted with a RealSense D435 RGB camera, which captures real-time RGB images at a resolution of 640 × 480 pixels and a frame rate of approximately 30 Hz.
To assess the zero-shot sim-to-real transfer capability of our VLA models, we directly applied the VLAs, trained using RGB images in the RoboTwin simulator, to three real-world manipulation tasks: Click Bell, Place Container Plate, and Pick Diverse Bottles. Each task is tested with 20
trials. Table \ref{tab:real_world} displays the evaluation results, where $\mathcal{M}^2$Tok attained an average success rate of 0.33 across all these tasks, consistently outperforming the baselines. This demonstrates that our $\mathcal{M}^2$Tok tokenizer can effectively adapt to real-world scenarios.
Additionally, Figure \ref{fig:real_robot} showcases two examples of manipulation processes performed by $\mathcal{M}^2$Tok. The top and bottom images illustrate the execution of the ``Place Container Plate'' and ``Pick Diverse Bottles'' tasks, respectively. These cases highlight the model’s ability to accurately identify the spatial positions of objects and execute precise grasping operations to complete tasks.

\section{Conclusion}
In this work, we have presented $\mathcal{M}^2$Tok, a novel action tokenization framework that fundamentally addresses the ``discretization bottleneck'' hindering current Vision-Language-Action (VLA) models. By departing from monolithic quantization strategies, our approach decomposes high-dimensional action signals into orthogonal subspaces and leverages the combinatorial density of multi-head codebooks. This architectural shift enables $\mathcal{M}^2$Tok to maximize representational expressivity, achieving a level of reconstruction fidelity previously unattainable by standard vector quantization or frequency-based methods. Across the RoboTwin and Simpler-Env benchmarks, as well as three challenging zero-shot real-world tasks, $\mathcal{M}^2$Tok-based VLA demonstrates significant improvements in task success rates compared to existing methods. Furthermore, comprehensive ablation studies confirmed that our structural innovations, specifically subspace decomposition and combinatorial codebook assignment, are the critical drivers of these gains. 

\noindent\textbf{Acknowledgement.}
This work was supported in part by the Shanghai Magnolia Talent Program Pujiang Project under Grant No. 25PJA076.

%
%
\bibliographystyle{splncs04}
\bibliography{main}

@article{liu2024deepseek,
  title={Deepseek-v3 technical report},
  author={Liu, Aixin and Feng, Bei and Xue, Bing and Wang, Bingxuan and Wu, Bochao and Lu, Chengda and Zhao, Chenggang and Deng, Chengqi and Zhang, Chenyu and Ruan, Chong and others},
  journal={arXiv preprint arXiv:2412.19437},
  year={2024}
}

@article{liang2025discrete,
  title={Discrete Diffusion VLA: Bringing Discrete Diffusion to Action Decoding in Vision-Language-Action Policies},
  author={Liang, Zhixuan and Li, Yizhuo and Yang, Tianshuo and Wu, Chengyue and Mao, Sitong and Pei, Liuao and Yang, Xiaokang and Pang, Jiangmiao and Mu, Yao and Luo, Ping},
  journal={arXiv preprint arXiv:2508.20072},
  year={2025}
}

@inproceedings{chi2023diffusion,
  title={Diffusion Policy: Visuomotor Policy Learning via Action Diffusion},
  author={Chi, Cheng and Feng, Siyuan and Du, Yilun and Xu, Zhenjia and Cousineau, Eric and Burchfiel, Benjamin and Song, Shuran},
  booktitle={Robotics: Science and Systems},
  year={2023}
}

@article{dubey2024llama,
  title={The llama 3 herd of models},
  author={Grattafiori, Aaron and Dubey, Abhimanyu and Jauhri, Abhinav and Pandey, Abhinav and Kadian, Abhishek and Al-Dahle, Ahmad and Letman, Aiesha and Mathur, Akhil and Schelten, Alan and Vaughan, Alex and others},
  journal={arXiv preprint arXiv:2407.21783},
  year={2024}
}

@article{achiam2023gpt,
  title={Gpt-4 technical report},
  author={Achiam, Josh and Adler, Steven and Agarwal, Sandhini and Ahmad, Lama and Akkaya, Ilge and Aleman, Florencia Leoni and Almeida, Diogo and Altenschmidt, Janko and Altman, Sam and Anadkat, Shyamal and others},
  journal={arXiv preprint arXiv:2303.08774},
  year={2023}
}

@article{tian2024visual,
  title={Visual autoregressive modeling: Scalable image generation via next-scale prediction},
  author={Tian, Keyu and Jiang, Yi and Yuan, Zehuan and Peng, Bingyue and Wang, Liwei},
  journal={Advances in neural information processing systems},
  volume={37},
  pages={84839--84865},
  year={2024}
}

@article{sun2024autoregressive,
  title={Autoregressive model beats diffusion: Llama for scalable image generation},
  author={Sun, Peize and Jiang, Yi and Chen, Shoufa and Zhang, Shilong and Peng, Bingyue and Luo, Ping and Yuan, Zehuan},
  journal={arXiv preprint arXiv:2406.06525},
  year={2024}
}

@inproceedings{
ji2025wavtokenizer,
title={WavTokenizer: an Efficient Acoustic Discrete Codec Tokenizer for Audio Language Modeling},
author={Shengpeng Ji and Ziyue Jiang and Wen Wang and Yifu Chen and Minghui Fang and Jialong Zuo and Qian Yang and Xize Cheng and Zehan Wang and Ruiqi Li and Ziang Zhang and Xiaoda Yang and Rongjie Huang and Yidi Jiang and Qian Chen and Siqi Zheng and Zhou Zhao},
booktitle={The Thirteenth International Conference on Learning Representations},
year={2025},
}

@inproceedings{
zhang2024speechtokenizer,
title={SpeechTokenizer: Unified Speech Tokenizer for Speech Language Models},
author={Xin Zhang and Dong Zhang and Shimin Li and Yaqian Zhou and Xipeng Qiu},
booktitle={The Twelfth International Conference on Learning Representations},
year={2024},
}

@article{brohan2022rt,
  title={Rt-1: Robotics transformer for real-world control at scale},
  author={Brohan, Anthony and Brown, Noah and Carbajal, Justice and Chebotar, Yevgen and Dabis, Joseph and Finn, Chelsea and Gopalakrishnan, Keerthana and Hausman, Karol and Herzog, Alex and Hsu, Jasmine and others},
  journal={arXiv preprint arXiv:2212.06817},
  year={2022}
}

@article{kim2024openvla,
  title={OpenVLA: An Open-Source Vision-Language-Action Model},
  author={Moo Jin Kim and Karl Pertsch and Siddharth Karamcheti and Ted Xiao and Ashwin Balakrishna and Suraj Nair and Rafael Rafailov and Ethan Foster and Grace Lam and Pannag R. Sanketi and Quan Vuong and Thomas Kollar and Benjamin Burchfiel and Russ Tedrake and Dorsa Sadigh and Sergey Levine and Percy Liang and Chelsea Finn},
  journal={ArXiv},
  year={2024},
  volume={abs/2406.09246},
}

@inproceedings{zitkovich2023rt,
  title={Rt-2: Vision-language-action models transfer web knowledge to robotic control},
  author={Zitkovich, Brianna and Yu, Tianhe and Xu, Sichun and Xu, Peng and Xiao, Ted and Xia, Fei and Wu, Jialin and Wohlhart, Paul and Welker, Stefan and Wahid, Ayzaan and others},
  booktitle={Conference on Robot Learning},
  pages={2165--2183},
  year={2023},
  organization={PMLR}
}

@article{Belkhale2024RTHAH,
  title={RT-H: Action Hierarchies Using Language},
  author={Suneel Belkhale and Tianli Ding and Ted Xiao and Pierre Sermanet and Quon Vuong and Jonathan Tompson and Yevgen Chebotar and Debidatta Dwibedi and Dorsa Sadigh},
  journal={ArXiv},
  year={2024},
  volume={abs/2403.01823},
}

@article{Team2024OctoAO,
  title={Octo: An Open-Source Generalist Robot Policy},
  author={Octo Model Team and Dibya Ghosh and Homer Rich Walke and Karl Pertsch and Kevin Black and Oier Mees and Sudeep Dasari and Joey Hejna and Tobias Kreiman and Charles Xu and Jianlan Luo and You Liang Tan and Pannag R. Sanketi and Quan Vuong and Ted Xiao and Dorsa Sadigh and Chelsea Finn and Sergey Levine},
  journal={ArXiv},
  year={2024},
  volume={abs/2405.12213},
}

@inproceedings{Seungjae2024vqbet,
  author={Seungjae Lee and Yibin Wang and Haritheja Etukuru and H. Jin Kim and Nur Muhammad Mahi Shafiullah and Lerrel Pinto},
  title={Behavior Generation with Latent Actions},
  year={2024},
  cdate={1704067200000},
  booktitle={ICML},
}

@article{pertsch2025fast,
  title={Fast: Efficient action tokenization for vision-language-action models},
  author={Pertsch, Karl and Stachowicz, Kyle and Ichter, Brian and Driess, Danny and Nair, Suraj and Vuong, Quan and Mees, Oier and Finn, Chelsea and Levine, Sergey},
  journal={arXiv preprint arXiv:2501.09747},
  year={2025}
}

@article{zhao2023ACT,
  title={Learning fine-grained bimanual manipulation with low-cost hardware},
  author={Zhao, Tony Z and Kumar, Vikash and Levine, Sergey and Finn, Chelsea},
  journal={arXiv preprint arXiv:2304.13705},
  year={2023}
}

@article{Wang2024ScalingPL,
  title={Scaling Proprioceptive-Visual Learning with Heterogeneous Pre-trained Transformers},
  author={Lirui Wang and Xinlei Chen and Jialiang Zhao and Kaiming He},
  journal={ArXiv},
  year={2024},
  volume={abs/2409.20537},
}

@article{Wu2023GR_1,
  title={Unleashing Large-Scale Video Generative Pre-training for Visual Robot Manipulation},
  author={Hongtao Wu and Ya Jing and Chi-Hou Cheang and Guangzeng Chen and Jiafeng Xu and Xinghang Li and Minghuan Liu and Hang Li and Tao Kong},
  journal={ArXiv},
  year={2023},
  volume={abs/2312.13139},
}

@article{Cheang2024GR_2,
  title={GR-2: A Generative Video-Language-Action Model with Web-Scale Knowledge for Robot Manipulation},
  author={Chi-Lam Cheang and Guangzeng Chen and Ya Jing and Tao Kong and Hang Li and Yifeng Li and Yuxiao Liu and Hongtao Wu and Jiafeng Xu and Yichu Yang and Hanbo Zhang and Minzhao Zhu},
  journal={ArXiv},
  year={2024},
  volume={abs/2410.06158},
}

@article{Zhai2023Siglip,
  title={Sigmoid Loss for Language Image Pre-Training},
  author={Xiaohua Zhai and Basil Mustafa and Alexander Kolesnikov and Lucas Beyer},
  journal={2023 IEEE/CVF International Conference on Computer Vision (ICCV)},
  year={2023},
  pages={11941-11952},
}

@article{gage1994new,
  title={A new algorithm for data compression},
  author={Gage, Philip},
  journal={C Users Journal},
  volume={12},
  number={2},
  pages={23--38},
  year={1994},
  publisher={McPherson, KS: R \& D Publications, c1987-1994.}
}

@misc{kim2025openvla_oft,
      title={Fine-Tuning Vision-Language-Action Models: Optimizing Speed and Success}, 
      author={Moo Jin Kim and Chelsea Finn and Percy Liang},
      year={2025},
      eprint={2502.19645},
      archivePrefix={arXiv},
      primaryClass={cs.RO},
}

@article{Chen2025RoboTwin,
  title={RoboTwin 2.0: A Scalable Data Generator and Benchmark with Strong Domain Randomization for Robust Bimanual Robotic Manipulation},
  author={Tianxing Chen and Zanxin Chen and Baijun Chen and Zijian Cai and Yibin Liu and Qiwei Liang and Zixuan Li and Xianliang Lin and Yiheng Ge and Zhenyu Gu and Weiliang Deng and Yubin Guo and Tian Nian and Xuanbing Xie and Qiangyu Chen and Kailun Su and Tianling Xu and Guodong Liu and Mengkang Hu and Huan-ang Gao and Kaixuan Wang and Zhixuan Liang and Yusen Qin and Xiaokang Yang and Ping Luo and Yao Mu},
  journal={ArXiv},
  year={2025},
  volume={abs/2506.18088},
}

@article{Shridhar2021CLIPortWA,
  title={CLIPort: What and Where Pathways for Robotic Manipulation},
  author={Mohit Shridhar and Lucas Manuelli and Dieter Fox},
  journal={ArXiv},
  year={2021},
  volume={abs/2109.12098},
}

@article{Shridhar2022PerceiverActorAM,
  title={Perceiver-Actor: A Multi-Task Transformer for Robotic Manipulation},
  author={Mohit, Shridhar and Lucas, Manuelli and Dieter, Fox},
  journal={ArXiv},
  year={2022},
  volume={abs/2209.05451},
}

@article{Intelligence2025pi_0_5,
  title={$\pi$0.5: a Vision-Language-Action Model with Open-World Generalization},
  author={Physical Intelligence and Kevin Black and Noah Brown and James Darpinian and Karan Dhabalia and Danny Driess and Adnan Esmail and Michael Equi and Chelsea Finn and Niccolo Fusai and Manuel Y. Galliker and Dibya Ghosh and Lachy Groom and Karol Hausman and Brian Ichter and Szymon Jakubczak and Tim Jones and Liyiming Ke and Devin LeBlanc and Sergey Levine and Adrian Li-Bell and Mohith Mothukuri and Suraj Nair and Karl Pertsch and Allen Z. Ren and Lucy Xiaoyang Shi and Laura Smith and Jost Tobias Springenberg and Kyle Stachowicz and James Tanner and Quan Vuong and Homer Rich Walke and Anna Walling and Haohuan Wang and Lili Yu and Ury Zhilinsky},
  journal={ArXiv},
  year={2025},
  volume={abs/2504.16054},
}

@article{Bu2025UniVLALT,
  title={UniVLA: Learning to Act Anywhere with Task-centric Latent Actions},
  author={Qingwen Bu and Yanting Yang and Jisong Cai and Shenyuan Gao and Guanghui Ren and Maoqing Yao and Ping Luo and Hongyang Li},
  journal={ArXiv},
  year={2025},
  volume={abs/2505.06111},
}

@article{Liu2025HybridVLA,
  title={HybridVLA: Collaborative Diffusion and Autoregression in a Unified Vision-Language-Action Model},
  author={Jiaming Liu and Hao Chen and Pengju An and Zhuoyang Liu and Renrui Zhang and Chenyang Gu and Xiaoqi Li and Ziyu Guo and Sixiang Chen and Mengzhen Liu and Chengkai Hou and Mengdi Zhao and KC alex Zhou and Pheng-Ann Heng and Shanghang Zhang},
  journal={ArXiv},
  year={2025},
  volume={abs/2503.10631},
}

@article{Li2025UnifiedVA,
  title={Unified Video Action Model},
  author={Shuang Li and Yihuai Gao and Dorsa Sadigh and Shuran Song},
  journal={ArXiv},
  year={2025},
  volume={abs/2503.00200},
}

@article{Liu2024RDT,
  title={RDT-1B: a Diffusion Foundation Model for Bimanual Manipulation},
  author={Songming Liu and Lingxuan Wu and Bangguo Li and Hengkai Tan and Huayu Chen and Zhengyi Wang and Ke Xu and Hang Su and Jun Zhu},
  journal={ArXiv},
  year={2024},
  volume={abs/2410.07864},
}

@article{Hu2024VideoPP,
  title={Video Prediction Policy: A Generalist Robot Policy with Predictive Visual Representations},
  author={Yucheng Hu and Yanjiang Guo and Pengchao Wang and Xiaoyu Chen and Yen-Jen Wang and Jianke Zhang and Koushil Sreenath and Chaochao Lu and Jianyu Chen},
  journal={ArXiv},
  year={2024},
  volume={abs/2412.14803},

}

@article{Nvidia2025GR00TNA,
  title={GR00T N1: An Open Foundation Model for Generalist Humanoid Robots},
  author={Nvidia and Johan Bjorck and Fernando Castaneda and Nikita Cherniadev and Xingye Da and Runyu Ding and LinxiJimFan and Yu Fang and Dieter Fox and Fengyuan Hu and Spencer Huang and Joel Jang and Zhenyuan Jiang and Jan Kautz and Kaushil Kundalia and Lawrence Lao and Zhiqi Li and Zongyu Lin and Kevin Lin and Guilin Liu and Edith Llontop and Loic Magne and Ajay Mandlekar and Avnish Narayan and Soroush Nasiriany and Scott Reed and You Liang Tan and Guanzhi Wang and Zu Wang and Jing Wang and Qi Wang and Jiannan Xiang and Yuqi Xie and Yinzhen Xu and Zhen-Teng Xu and Seonghyeon Ye and Zhiding Yu and Ao Zhang and Hao Zhang and Yizhou Zhao and Ruijie Zheng and Yuke Zhu},
  journal={ArXiv},
  year={2025},
  volume={abs/2503.14734},
}

@article{van2017neural,
  title={Neural discrete representation learning},
  author={Van Den Oord, Aaron and Vinyals, Oriol and others},
  journal={Advances in neural information processing systems},
  volume={30},
  year={2017}
}

@article{Yang2024Qwen2_5,
  title={Qwen2.5 Technical Report},
  author={Qwen An Yang and Baosong Yang and Beichen Zhang and Binyuan Hui and Bo Zheng and Bowen Yu and Chengyuan Li and Dayiheng Liu and Fei Huang and Guanting Dong and Haoran Wei and Huan Lin and Jian Yang and Jianhong Tu and Jianwei Zhang and Jianxin Yang and Jiaxin Yang and Jingren Zhou and Junyang Lin and Yang Su and Yi-Chao Zhang and Yunyang Wan and Yuqi Liu and Zeyu Cui and Zhenru Zhang and Zihan Qiu and Shanghaoran Quan and Zekun Wang},
  journal={ArXiv},
  year={2024},
  volume={abs/2412.15115},
}

@inproceedings{Loshchilov2017DecoupledWD,
  title={Decoupled Weight Decay Regularization},
  author={Ilya Loshchilov and Frank Hutter},
  booktitle={International Conference on Learning Representations},
  year={2017},
}

@article{Wang2025VQVLA,
  title={VQ-VLA: Improving Vision-Language-Action Models via Scaling Vector-Quantized Action Tokenizers},
  author={Yating Wang and Haoyi Zhu and Mingyu Liu and Jiange Yang and Hao-Shu Fang and Tong He},
  journal={ArXiv},
  year={2025},
  volume={abs/2507.01016},
  url={https://api.semanticscholar.org/CorpusID:280145409}
}

@inproceedings{walke2023bridgedata,
  title={Bridgedata v2: A dataset for robot learning at scale},
  author={Walke, Homer Rich and Black, Kevin and Zhao, Tony Z and Vuong, Quan and Zheng, Chongyi and Hansen-Estruch, Philippe and He, Andre Wang and Myers, Vivek and Kim, Moo Jin and Du, Max and others},
  booktitle={Conference on Robot Learning},
  pages={1723--1736},
  year={2023},
  organization={PMLR}
}

@inproceedings{DBLP:conf/nips/MaJWYYYPQ25,
  author       = {Chuofan Ma and
                  Yi Jiang and
                  Junfeng Wu and
                  Jihan Yang and
                  Xin Yu and
                  Zehuan Yuan and
                  Bingyue Peng and
                  Xiaojuan Qi},
  editor       = {Danielle Belgrave and
                  Cheng Zhang and
                  Laura N. Montoya and
                  Hsuan{-}Tien Lin and
                  Razvan Pascanu and
                  Piotr Koniusz and
                  Marzyeh Ghassemi and
                  Nancy Chen and
                  Iv{\'{a}}n Vladimir Meza Ru{\'{\i}}z and
                  Arturo Loaiza{-}Bonilla},
  title        = {UniTok: a Unified Tokenizer for Visual Generation and Understanding},
  booktitle    = {Advances in Neural Information Processing Systems 38: Annual Conference
                  on Neural Information Processing Systems 2025, NeurIPS 2025, San Diago,
                  CA, USA, December 2-7, 2025 / Mexico City, Mexico, November 30 - December
                  5, 2025},
  year         = {2025},
  url          = {http://papers.nips.cc/paper\_files/paper/2025/hash/bbe04d329d531e8814f1199098bd8fb6-Abstract-Conference.html},
  bibsource    = {dblp computer science bibliography, https://dblp.org}
}
\end{document}